\documentclass[two column]{article}
\usepackage[]{algorithm2e}
\usepackage{amsmath}

\usepackage{graphicx}
\usepackage{amsfonts}
\usepackage{amssymb}
\usepackage{amsthm}
\usepackage{subfig}
\usepackage{rotating}
\usepackage{threeparttable}

\newlength{\tempheight}
\newlength{\tempwidth}
\newcommand{\rowname}[1]% #1 = text
{\rotatebox{90}{\makebox[\tempheight][c]{#1}}}

\newcommand{\columnname}[1]% #1 = text
{\makebox[\tempwidth][c]{#1}}

\newcommand{\ua}{\uparrow}
\newcommand{\nc}{\newcommand}
\nc{\da}{\downarrow} \nc{\hc}{\hat{c}} \nc{\hS}{\hat{S}}
\nc{\bra}{\langle} \nc{\ket}{\rangle} \nc{\eq}{equation (\ref}
\nc{\h}{\hat} \nc{\hT}{\h{T}}\nc{\be}{\begin{eqnarray}}
\nc{\ee}{\end{eqnarray}}\nc{\rd}{\textrm{d}}\nc{\e}{eqnarray}\nc{\hR}{\hat{R}}\nc{\Tr}{\mathrm{Tr}}
\nc{\tS}{\tilde{S}}\nc{\tr}{\mathrm{tr}}\nc{\8}{\infty}\nc{\lgs}{\bra\ua,\phi|}\nc{\rgs}{|\ua,\phi\ket}
\nc{\hU}{\hat{U}}\nc{\lfs}{\bra\phi|}\nc{\rfs}{|\phi\ket}\nc{\hZ}{\hat{Z}}\nc{\hd}{\hat{d}}\nc{\mD}{\mathcal{D}}
\nc{\bd}{\bar{d}}\nc{\bc}{\bar{c}}\nc{\mc}{\mathcal}\nc{\ea}{eqnarray}\nc{\mG}{\mathcal{G}}\nc{\bce}{\begin{center}}
\nc{\ece}{\end{center}}

\newcommand{\x}{x}

\renewcommand{\xi}{{\x^i}}

\begin{document}

\title{Spatially Aware Melanoma Segmentation Using Hybrid Deep Learning Techniques}
\author{M.~Attia$^{\star}$, M.~Hossny$^{\star}$, S.~Nahavandi$^{\star}$ and A.~Yazdabadi$^{\dagger}$\\ $^{\star}$ Institute for Intelligent Systems Research and Innovation, Deakin University \\ $^{\dagger}$ School of Medicine, Deakin University}

\maketitle

\begin{abstract}
In this paper, we proposed using a hybrid method that utilises deep convolutional and recurrent neural networks for accurate delineation of skin lesion of images supplied with ISBI 2017 lesion segmentation challenge. The proposed method was trained using 1800 images and tested on 150 images from ISBI 2017 challenge. 
\end{abstract}

%\begin{keyword}
%Heteroskedasticity, Bhattacharyya distance, KL-Divergence
%%\texttt{elsarticle.cls}\sep \LaTeX\sep Elsevier \sep template
%%\MSC[2010] 00-01\sep  99-00
%\end{keyword}

%\end{frontmatter}

\section{Introduction}
\label{sec:intro}
\smallbreak
Melanoma is one of the most deadliest types of cancer that affects large sector of population in United States and Australia. It is responsible for more than 10,000 deaths in 2016. Clinicians diagnose melanoma by visual inspection of skin lesions and moles~\cite{doukas2015skin}. In this work, we propose an novel approach to segment lesions using deep neural networks. We compared our results to popular deep learning semantic segmentation convolutional neural networks FCN ~\cite{long2015fully} and SegNet~\cite{badrinarayanan2015segnet}. This approach will be presented in the in International Symposium on Biomedical Imaging 2017. 

\smallbreak
The rest of this paper is organised as follows. Section 2 describes the related work. The proposed method is presented in Section 3. Section 4 presents results and, finally, Section 5 concludes. 

%\section{Lesion Segmentation Challenges}

\section{Related Work}
\label{sec:related}
Traditional intensity based segmentations achieved high accuracies. However, low contrast images with high variance uni-modal histograms resulted in inaccurate delineation of borders. Most of these inaccuracies were corrected with post-processing of images~\cite{EmreCelebi2007, Xie2013,attia2016}. 

Deep convolutional neural network (CNN) with auto encoder-decoder architectures achieved great results in semantic segmentation~\cite{long2015fully}. Upsampling methods were proposed to solve lost spatial resolution ~\cite{long2015fully}. Ronneberger et al. concatenated a copy of encoded feature map during decoding phase to increase spatial accuracy of the output feature maps ~\cite{ronneberger2015u}. Zheng et al. proposed a trainable conditional random field (CRF) module to refine segmentation prediction map~\cite{zheng2015conditional}. Visin et al. proposed a recurrent neural network (RNN) as post processing module for the coarse extracted feature maps~\cite{visin2015reseg}.
%In the following section, we will define a network based on the proposed method by Visin et al. for lesion segmentation~\cite{visin2015reseg}. This network used a joint scheme of convolutional and recurrent neural networks.

\begin{figure*}[t]
	\centering
	\includegraphics[width=1\linewidth]{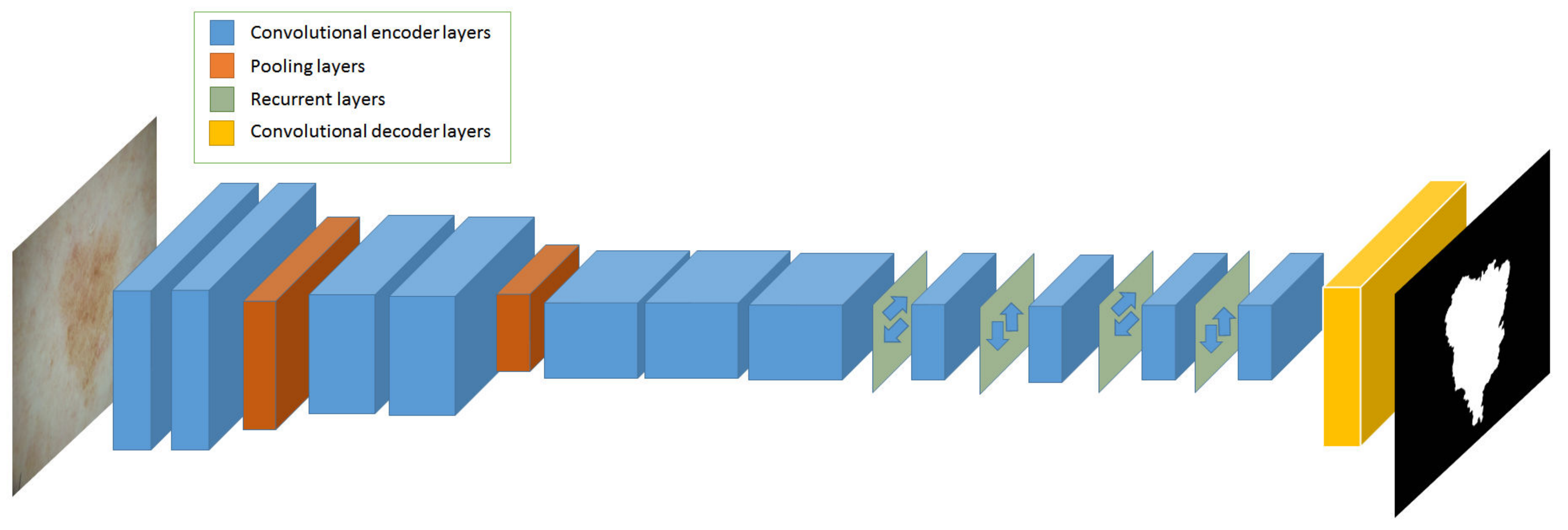}
	\caption{Proposed architecture for RNN and CNN. Auto encoder network consists of 7-convolutional layers with 2 max-pooling layers. Then, extracted feature maps are fed into 4 layers of recurrent network with 4 decoupled direction. The mask is reconstructed using auto decoder network \cite{attia2017}.}
	\label{fig:network}
\end{figure*}

\section{Proposed Hybrid Deep Architecture}
\label{sec:METHODS}
The main drawback of semantic segmentation with fully convolutional neural networks (FCN ~\cite{long2015fully} and SegNet~\cite{badrinarayanan2015segnet}) is over segmentation due to coarse output of the max-pooling layers during the encoding phase. In order to address this problem, we propose to use recurrent neural networks to learn the spatial dependencies between active neurones after the max-pool encoding~\cite{visin2015reseg}.

The RNN layers are fed with flattened non-overlapping data patches to model spatial dependencies. Let  $D$  is the input data such that $D  \hspace{2pt} \in \hspace{2pt} \mathbb{R}^{w \times h \times c}$ where $ w,h $ and $ c $ are width, height and channels respectively. $D$ is splitted into $ n \times m $ patches $P_{i,j}$ such that $P_{i,j} \hspace{2pt}  \in \hspace{2pt} \mathbb{R}^{w_p\times h_p \times c}$ where $w_p=w/n$ \hspace{2pt}and \hspace{2pt}$h_p=h/m$. Input patches are flattened into 1-D vector to update its hidden state $z^*_{i,j}$ where $*$ is the direction of the sweep direction $\uparrow,\downarrow,\rightarrow$ and $\leftarrow$.

\smallbreak
For every patch $P_{i,j}$, the composite activation map feature $ O =\{o^*_{i,j}\}_{\{i=1,2,\dots,n\}}^{\{j=1,2,\dots,m\}}$ is concatenation of output activation two coupled direction RNN either horizontal (right to left and left to right) or vertical sweep (up to down and down to up) where $ o^*_{i,j} \in \mathbb{R}^{2U} \hspace{1pt} \forall \hspace{1pt} * \in \{(\uparrow,\downarrow),(\rightarrow,\leftarrow)\} $ is activation of the recurrent unit at position $ ({i,j}) $ with respect to all patches in the column $j$ in case of coupled vertical direction $\{(\downarrow,\uparrow)\}$ and to all patches in the row $i$ in case of coupled horizontal sweep $\{(\rightarrow,\leftarrow)\}$ and $O^\updownarrow$ denotes concatenated output of $o^\downarrow$ and $ o^\uparrow $ and similarly $O^\leftrightarrow$ for $O^\leftarrow$ and $O^\rightarrow$ and $U$ is the number of recurrent units.

Similarly, $o^\downarrow_{i,j}$ and coupled horizontal sweep function can be defined. It is worth noting that both directions are computed independently.

\smallbreak
Finally, in the decoding stage, deeply encoded features by sequenced recurrent units are used to reconstruct the segmentation mask at the same resolution of the input. Fractionally strided convolutions were used in reconstruction of final output. In strided convolutions, prediction are calculated by inner product between the flattened input and a sparse matrix, whose non-zero elements are elements of the convolutional
kernel. This method is both computationally and memory efficient to support joint training of convolutional and recurrent neural networks~\cite{dumoulin2016guide}.

\section{Results}
\label{sec:EXPERIMENTS}
\begin{figure}[t]
	\begin{minipage}[b]{1\linewidth}
		\centering
		\centerline{\includegraphics[width=0.99\linewidth]{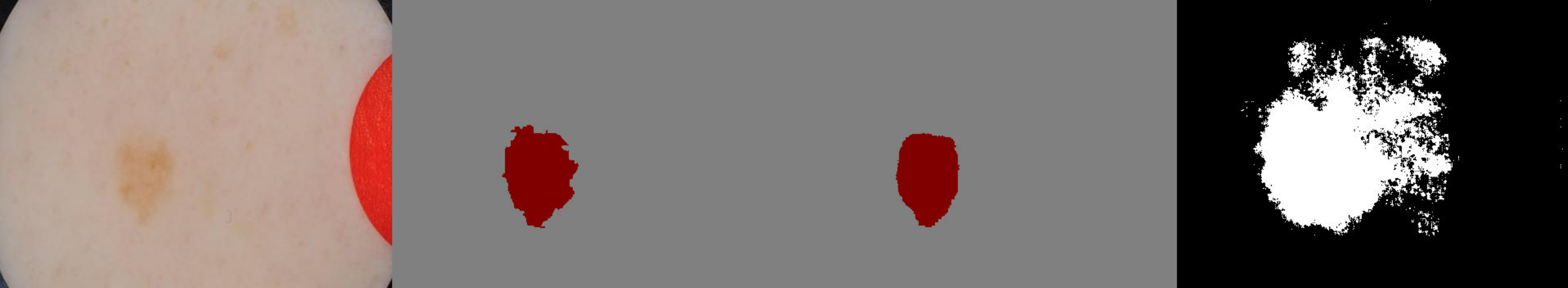}}
		%  \vspace{1.5cm}
		\centerline{(a) Segmentation of a low contrast lesion}\medskip
	\end{minipage}
	\begin{minipage}[b]{1\linewidth}
		\centering
		\centerline{\includegraphics[width=0.99\linewidth]{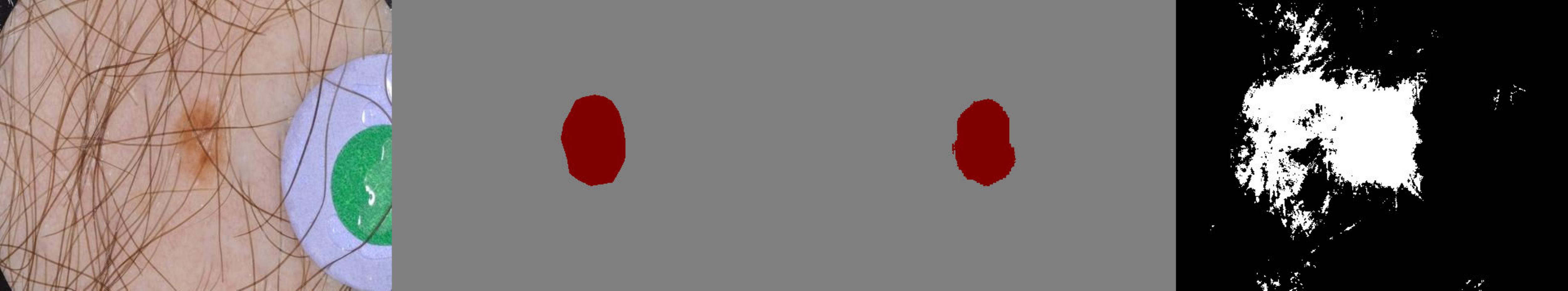}}
		%  \vspace{1.5cm}
		\centerline{(b) Segmentation result of a hair occluded lesion}\medskip
	\end{minipage}
	\caption{Example of lesions segmentation. From left to right: image, ground truth, proposed method and SegNet. The output of proposed method does not require any contrast enhancement or postprocessing operations compared to SegNet.}	\label{fig:res}
\end{figure}

The proposed network was trained using 1800 lesion images provided along with ground truth. These images were provided for the first task of ISBI 2017 challenge ``Skin Lesion Analysis Toward Melanoma Detection''\cite{GutmanCCHMMH16}. The performance of the proposed method is compared to other methods using pixel-wise metrics: Jaccard index, accuracy, sensitivity, specificity and dice coefficient. The results shown in Fig.~\ref{fig:res} demonstrate the efficacy of the proposed method compared to over the classical SegNet \cite{attia2017}. These results were obtained on the ISBI training dataset released in January, 2017. The results tabulated in Table~\ref{table:results} will be presented in ISBI 2017 \cite{attia2017}.

\begin{figure}[t]
	\begin{minipage}[b]{1\linewidth}
		\centering
		\includegraphics[width=0.99\linewidth]{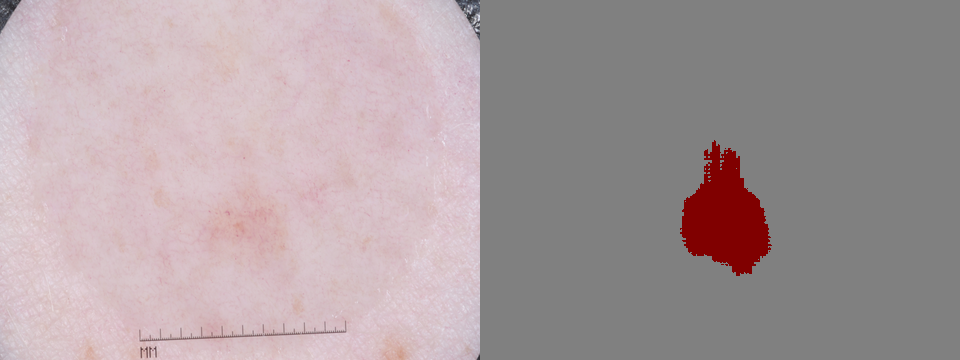}
		\centerline{(a) Low contrast lesion}\medskip
	\end{minipage}
	\begin{minipage}[b]{1\linewidth}
		\centering
		\centerline{\includegraphics[width=0.99\linewidth]{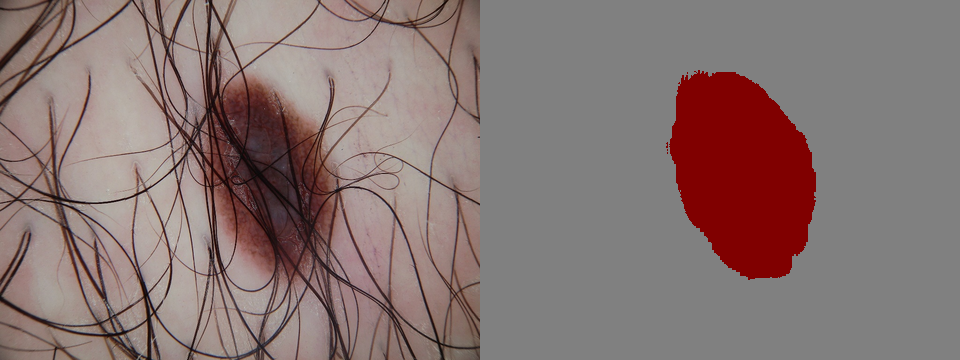}}
		%  \vspace{1.5cm}
		\centerline{(b) Hair occluded lesion}\medskip
	\end{minipage}
	\caption{Samples of the output segmentation mask of validation set. Ground truth masks are not released yet.}	\label{fig:res17}
\end{figure}

\begin{figure}[t]
	\begin{minipage}[b]{1\linewidth}
		\centering
		\includegraphics[width=0.99\linewidth]{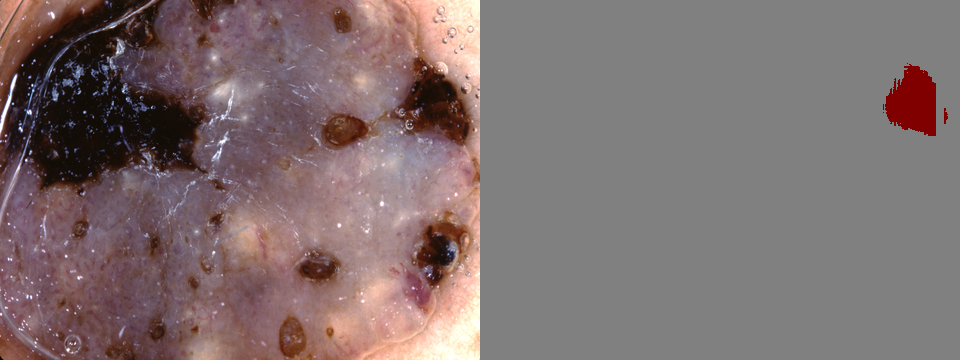}
		\centerline{(a) Compound lesions with different colours}\medskip
	\end{minipage}
	\begin{minipage}[b]{1\linewidth}
		\centering
		\centerline{\includegraphics[width=0.99\linewidth]{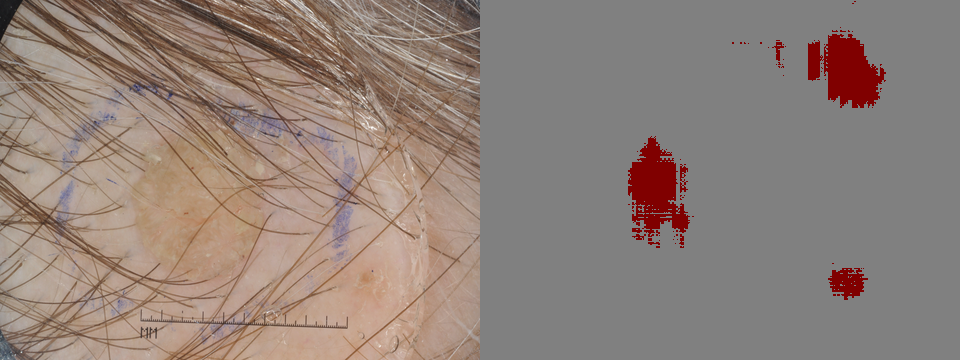}}
		%  \vspace{1.5cm}
		\centerline{(b) Hair occlusions and marker artifacts}\medskip
	\end{minipage}
	\caption{Samples of the test dataset with what is believed to be a bad segmentation mask. Ground truth masks are not released yet.}	\label{fig:res18}
\end{figure}

\smallbreak
Figure~\ref{fig:res17} and  Figure~\ref{fig:res18} show sample of the output masks. the ground truth are not published yet.

\begin{table}[b]
\caption{Lesion Segmentation Results. Higher results is better.}
\label{table:results}
	\begin{center} 
		\begin{tabular}{c c c c c c  }
			\hline 
			& AC & SE &SP&DI & JA\\ 
			\hline 
			SegNet~\cite{badrinarayanan2015segnet} & 0.91 &0.87&\textbf{ 0.96}
			&0.92 & 0.86 \\ 
			Proposed & \textbf{0.98} &\textbf{0.954}
			& 0.94& \textbf{0.96} & \textbf{0.93} \\
			FCN~\cite{long2015fully} &0.82
			& 0.85 & 0.70&
			0.82 &
			0.86
			\\ 
			
			\hline 
		\end{tabular} 
	\end{center}
\end{table}

\section{Conclusion}
\label{sec:CONCLUSION}
We utilised a joint architecture that incorporates both deep convolutional and recurrent neural networks for skin lesion segmentation. The results presented great potentials by outperforming state-of-the-art methods of segmentation on skin melanoma delineation problem. Also, it is immune, with high sensitivity, to all artifacts such as markers, ruler marks, and hair occlusions.
\smallbreak
\noindent

\section*{Acknowledgement}
This research was fully supported by the Institute for Intelligent Systems Research and Innovation (IISRI).

%\section*{References}
%\bibliographystyle{els/model2-names}
%\printbibliography

\bibliography{QH15}

\end{document}